\documentclass[11pt,times]{article}
\usepackage{multirow}
\usepackage{fullpage}
\usepackage{lipsum}
\usepackage{amsthm}
\usepackage{listings}
\usepackage{wrapfig}
\usepackage{amsmath}
\usepackage{booktabs}
\usepackage{graphicx}
\usepackage{xspace}
\usepackage{colortbl}
\usepackage{tabularx}
\usepackage{csquotes}
\usepackage{times} 
\usepackage{array}
\usepackage{tcolorbox}
\usepackage{multirow}
\usepackage{adjustbox}
\usepackage{rotating}
\usepackage{enumitem}

\title{Software Engineering for Robotics: Future Research Directions
\\ \vspace{0.1in} \Large{Report from the 2023 Workshop on Software Engineering for Robotics}}
\author{Claire Le Goues, Sebastian Elbaum, David Anthony, Celik Berkay, Mauricio Castillo-Effen,  \\ Nikolaus Correll, Pooyan Jamshidi,    Morgan Quigley,   Trenton Tabor, and 
Qi Zhu
 }
 
\begin{document}

\maketitle

\vspace{-0.5in}
\section*{Executive Summary}

Robots are experiencing a revolution as they permeate many aspects of our daily lives, from performing house maintenance to infrastructure inspection, from efficiently warehousing goods to autonomous vehicles, and more. This technical progress and its impact are astounding. 
This revolution, however, is outstripping the capabilities of existing software development processes, techniques, and tools, which largely have remained unchanged for decades. These capabilities are ill-suited to handling the challenges unique to robotics software such as dealing with a 
wide diversity of domains, heterogeneous hardware, programmed and learned components, complex physical environments  captured and modeled with uncertainty, emergent behaviors that include human interactions, and scalability demands that span across multiple dimensions.   

Looking ahead to the need to develop software for robots that are ever more ubiquitous, autonomous, and reliant on complex adaptive components, hardware, and data, motivated an NSF-sponsored community workshop on the subject of Software Engineering for Robotics, held in Detroit, Michigan in October 2023.  The goal of the workshop was to bring together thought leaders across robotics and software engineering to coalesce a community, and identify key problems in the area of SE for robotics that that community should aim to solve over the next 5 years. This report serves to summarize the motivation, activities, and findings of that workshop, in particular by articulating the challenges unique to robot software, and identifying a vision for fruitful near-term research directions to tackle them:

\begin{itemize}[leftmargin=0.15in]
\setlength\itemsep{0.05em}
\item \textbf{Massive Simulation Eco-systems} to collaboratively and systematically challenge robots from multiple simulators across complex simulated environments. 
\item \textbf{Curricula to Prepare Future SE-Robotics developers} in the foundational skills at the intersection of robotics and software engineering.
\item \textbf{Multi-view and full stack architectural design languages for robotics} that span multiple layers and views to overcome the gaps between areas of expertise that characterize development teams, and provide automated support to reduce integration faults and development costs. 
\item \textbf{Processes, Models, Estimators and Analyses for Humans and Robot Interactions} to guide both developers of robotics systems, and those systems themselves, in service of safe, complex deployments.

\item \textbf{Evidence-based processes and assurances for robotics development} to systematize knowledge and provide a solid basis for assurance across industry and practice. 

\item \textbf{Modern abstractions for resources into middlewares and libraries} that can handle the increased complexity of designing, configuring, and deploying state-of-the-art robotics systems.

\item \textbf{Multidimensional Quality Assurance} that caters to the needs of the robotics domain, handling component heterogeneity and interaction,  accounting for the complexity of the configuration spaces and adversarial threats, and the richness and uncertainties of the environments. 
\end{itemize}

\subsection*{Workshop Organization}

\begin{itemize}

\item Chairs:

\begin{itemize}
    \item  Claire Le Goues, Associate Professor of Computer Science, Carnegie Mellon University
    \item Sebastian Elbaum, Anita Jones Professor of Computer Science, University of Virginia
\end{itemize}

\item Steering Committee: 
\begin{itemize}
\item Bill Smart, Professor of Mechanical Engineering and Robotics, Oregon State University
\item Brian Gerkey, CEO and Co-Founder, Open Robotics 
\item David Wettergreen, Research Professor at the Robotics Institute, Carnegie Mellon University
\item Xiangyu Zhang, Professor of Computer Science, Purdue University
\end{itemize}

\item NSF Liaison:  Sol Greenspan

\item Contributing Participants
\label{sec:participants}

\begin{table} [h]
\footnotesize
\begin{tabular}{l l l l }
\hline
Name & Organization & Name & Organization \\
\hline
Claire Le Goues,  Chair & Carnegie Mellon University   & & \\					
Sebastian Elbaum,  Chair  & University of Virginia   &		Mauricio Castillo & Lockheed Martin \\			
Alfred Chen	& University California Irvine &	 Mona Rahimi	& University of Northern Illinois \\		
Anastasia Mavridou& NASA & 		Morgan Quigley	& Open Robotics \\		
Baishaki Ray & University of Columbia &		Morteza Lahijanian	& University of Colorado \\		
Betty Cheng	& Michigan State University &	Myra Cohen	& Iowa State University \\		
Bill Drozd & 	Carnegie Mellon University &	Nathan Brooks& Picknik \\			
Bradley Schmerl	& Schmerl Consulting &	Neno Medvidovic& University of Southern California \\			
Brian Williams	& Massachusetts Institute of Technology &	Nicholas Gans & University of Texas Arlington \\			
Christopher Timperley	& Carnegie Mellon University &	Nikolaus Correll & University of Colorado \\			
Chuchu Fan	& Massachusetts Institute of Technology &	Payam Ghassemi	& Caterpillar \\		
Chuck Claunch	&  Nauticus Robotics &	Pooyan Jamshidi	& University of South Carolina \\		
 Dave Anthony &	Southwest Research Institute &	Qi Zhu	& Northwestern University \\		
Dave Guttendorf	& Carnegie Mellon University &	Reid Simmons & 	Carnegie Mellon University \\		
David Wettergreen	& Carnegie Mellon University &	Richard M. Voyles	& Purdue University \\		
Eric Feron	& Georgia Tech and Kaust University &	Sabrina Neuman	& Boston University \\		
Erik Fredericks & 	Grand Valley State University &	Sanjit Seshia	& University California Berkeley \\		
Geoffrey Biggs	& Open Robotics & 	Somil Bansal &	University of Southern California \\		
Jane Clelang Huang	& University of Notre Dame &	Tichakorn  Wongpiromsarn	& Iowa State University \\		
Jeff Ichnowski	& Carnegie Mellon University &	Tingting Yu	& University of Connecticut \\		
John Dolan	& Carnegie Mellon University &	Todd Pack 	& Neocybernetica \\		
John-Paul Ore	& North Carolina State University &	Trenton Tabor	& 	Carnegie Mellon University \\	
Joshua Garcia	& University California Irvine &	Trey Woodlief	& University of Virginia   \\		
Katherine   Scott	& Open Robotics &	Wil Thomason  	&  Rice University \\		
Louise Poubel	 & Zipline &	Xiangyu Zhang	& Purdue University \\		
Madhur Behl	& University of Virginia &	Berkay Celik	& Purdue University \\		
Marsha Chechik &	University of Toronto &	Zak Kingston		&  Rice University \\	
\end{tabular} 
\end{table}

\end{itemize}

\subsection*{Workshop Support}

The workshop was made possible by the National Science Foundation through grant \#2332991.

\newpage
\section{Introduction}

Robots are experiencing a revolution as they permeate many aspects of our daily lives, 
from performing house maintenance to infrastructure inspection,  from efficiently warehousing goods to autonomous vehicles, from acting as museum hosts to carrying out safety and rescue missions, from picking up produce to conducting remote surgeries. This technical progress and its impact are astounding. Meanwhile, research initiatives in robotics around the world have surged.  Such research efforts combined with an emerging market have energized the robotics industry. 

Yet, software development techniques and tools have not kept up with this revolution, and in many cases are hindering it. With a few exceptions (e.g., the ROS framework, specialized libraries, targeted verification efforts), the software development processes, techniques, and tools employed today in robotics are the same as those first defined decades ago.
The fundamental challenges imposed by the original robotics systems (e.g., rich states governed in part by physical and temporal constraints, highly distributed, susceptibility to noise, every line of code has uncertainty, imbued with safety critical concerns) remain substantial. 
Those challenges are confounded and further amplified by today's diversity of
(1) robots, scenarios, and clients being deployed, (2) inputs they take and modern technologies and hardware they employ, and (3) the diversity of development environments, sources of uncertainty, and engineer backgrounds.  
Handling this diversity requires a dramatic shift in the development of the robot software stack. Looking ahead to robots that are more ubiquitous, autonomous, and reliant on complex adaptive components, hardware, and data makes the deficiencies of tools and techniques for robotics software development ever more pressing. 

To begin addressing how to tackle these challenges, we organized a 2-day workshop that brought together thought leaders from academia and industry, in robotics and software engineering, to identify key problems in software engineering for robotics that we should tackle in the next 5 years, and coalesce a community around those problems.
Of our 53 attendees, 5\% represented government agencies, 23\% were industry practitioners, and 72\% represented academic researchers; these participants' research interests were distributed approximately 56\%-44\% robotics-software engineering (more detail on the attendees is available in Section~\ref{sec:participants}). 
The workshop also generated new connections, collaborations, and synergies amongst the researchers in attendance. As stated by two senior leaders of each community: \textit{``Thank you again for arranging such a great workshop! What a whirlwind of activities and meeting new people''} 
and \textit{``Thank you very much for the invitation to your workshop.  It was quite inspirational.  I’ve attended lots of program formulation
meetings.  Your process was the most effective one I’ve participated in. ''} 

The workshop tapped into the diverse experiences and expertise of our participants to generate and synthesize the problems, vision, ideas, and research agenda that are summarized in this document (a detailed workshop agenda is available in Section~\ref{sec:agenda}).
In brief, the workshop included four framing talks to highlight different dimensions of the problem and the solution space.
However, most of the workshop schedule was dedicated to hands-on activities that: (1) identified  the most distinctive current and emerging software engineering practices in the robotics domain that require specialized support, and (2) determine the major software engineering pain points in robotics and how those are expected to shift in the future to require transformational software engineering techniques.  

The rest of the report discusses the dominant challenges for developing software for robotics and the proposed research directions to address those challenges identified during the workshop. Table \ref{tab:mapping} summarizes the mapping of challenges  to research directions, which are later elaborated to constitute a roadmap for the research community.

\newcommand\rot[1]{\begin{turn}{90}#1\enspace\end{turn}}

\begin{table}[ht!]
    \small
    \centering
    \begin{tabular}{lccccccccc}
         \textbf{Research Directions} & \rot{\textbf{Challenges}} & 
         \rot{Deployment in Physical Environment} & 
         \rot{HRI and Emerging Behaviors} &  
         \rot{Simulation Reality Gap} &
           \rot{Layered Leaky Heterogeneity}  & 
           \rot{Integration Learned Components} &
         \rot{Diverse Siloed Development} &   
          \rot{Broken Scalability Tenets in   Pipelines} \\ 
         \hline
         
         Modern Abstractions for Resources   & & $\times$&   & &$\times$ &   &  & $\times$ \\
         Multi-View and Full-Stack ADLs for Robotics &    &  & $\times$   &  & $\times$ & $\times$ & $\times$ &$\times$ \\
         Models, Estimators, and Analyses for HRI & &  $\times$  & $\times$ & $\times$ &  &  & \\
        Massive Simulation Eco-Systems & & $\times$  & $\times$ & $\times$   & &  & $\times$ &  \\
         Multi-Dimensional Quality Assurance for Robots & & $\times$ & $\times$ & $\times$ &   & $\times$ &   \\
          Evidence-based Processes and Assurances & &  &   & &  $\times$  & $\times$ &  $\times$ & $\times$\\
        Curricula to Prepare future SE-Robotics Developers & & $\times$  & $\times$ & $\times$ & $\times$ & $\times$ & $\times$ &  $\times$ \\
    \\ 

    \end{tabular}
    \caption{Research directions to address impeding challenges  }
    \label{tab:mapping}
\end{table}

\newpage
\section{Challenges}
 
Robotic systems have unique properties as software systems and as engineered artifacts.  These properties, along with the ecosystem of robotic researchers and practitioners, pose unique challenges to effective software engineering in the robotic context. We describe the most significant  emerging challenges identified in the workshop in this section.

\subsection{Deployment in a Physical Environment}

Robotics systems are ultimately deployed in and interact with physical environments.  
This introduces a wide range of factors that challenge the effective conception, development, and assurance of robotics software.  
First, the physical world affects robotic behavior in ways that traditional software design and analysis is ill-prepared to reason about, such as noisy sources of data coming from sensors and uncertain actuator dynamics, different frames of reference that must be coordinated, and performance variability that is difficult to model due to physical factors like manufacturing variability and tear-and-wear. 
Second, the physical environments are wildly diverse
and are often difficult-to-impossible to fully model and predict, let alone specify. A system's operational design domain can be a key consideration in development, but anticipating the distribution of relevant features of diverse environments is plagued with uncertainty,  and modeling non-linear phenomena such as friction and slip during contact remain hard problems.  
Third, deployment in the physical environment introduces key sources of non-determinism, at every level, from the environment itself, to the actuation effects on the environment. Finally, as a result of these factors and as discussed under the simulation-reality gap, it is difficult to construct simulators with sufficient fidelity to validate robotic systems. Traditional software engineering methods are not well-equipped to deal with these challenges, especially when occurring at scale.

\subsection{Human Interaction and other Emergent Dynamic Behavior} 

Robotics systems are increasingly designed for daily-life tasks, involving interaction with other dynamic agents and, critically, humans.  
Any notion of robotic software engineering correctness must account for interaction with such dynamic actors, and for emergent behaviors that arise over long deployments in unpredictable contexts.
Before robots are fully accepted as a part of daily tasks, they must act in a way that humans can trust, understand, predict, and accept.  To the extent that robots do not meet these expectations (or worse, cause accidents), humans will not accept their presence.  
Likewise, robots must be able to model and predict human actions and reactions in joint environments. 
Although models of human behavior, motion, belief, desire, and intention are available, unifying and composing these models for use by robotics developers remains an unsolved problem.  
Last, robots must possess a sufficient level of autonomy to responsibly adapt to those emerging behaviors, and due so by extracting uncertain and limited information from large amounts of data, while accounting for the constraints of the physical deployment environment and the robot resource capabilities.

\subsection{Simulation Reality Gap}
 
Simulation platforms provide virtual environments in which robots and their components can operate. 
These virtual environments offer the flexibility to more quickly and cheaply explore design alternatives, and validate components up to whole systems before deployment in the real physical world.
The fidelity of those virtual environments varies according to the simulation objective, required accuracy, and cost.
Fast low-fidelity simulators use mathematical models to approximate the world and the robot states; high-fidelity simulators include sophisticated physics and graphical engines.
Yet, despite their broad adoption in both industry and as part of academic curricula, and their cost-effectiveness, the virtual environments provided by simulators for robotics are still mere approximations of the real world. The difference between the real and the simulated worlds is known as the simulation reality gap. This gap explains, at least in part, why faults are detected in the field even after extensive simulation, why faults found in the simulation are not likely to occur in the field, and why field testing is still the gold standard for robotics. Furthermore, as robots are used in more extreme environments, such as space or severe disaster areas, the lack of ground truth data makes it challenging to develop and validate simulators and understand their gaps. Efforts targeting gap reduction through more sophisticated physics and graphics engines help, especially within a limited domain, but they do not generalize and currently seem insufficient to address the broad challenge.

\subsection{Layered, Leaky Heterogeneity}

Robotics systems are multi-layered and heterogeneous, composed of diverse software and hardware components, and a mix of programmed and machine-learned components (discussed in the next section).
Hardware and lower-level systems can fail in specific ways. However, these failures may not be exposed or may fail to propagate, due to contracts between layers that are difficult to maintain over the course of integration.   
Learned components are composed with programmed components, but they often have opaque internal states that are difficult to reason about in the composition. 
These layers and components also tend to be leaky; that is, their behavior changes in ways that are not captured by their documentation and specifications. 
Robot systems are also
typically structured as distributed systems facilitated in many cases by sophisticated middleware. This middleware is managed through a large number of multi-dimensional, inter-dependent variables and a mix of static and dynamic parameters scattered throughout source code, configuration, and deployment files.
This complex distributed space of parameters and layers makes it very difficult to effectively compose, analyze, debug, and maintain robotic software.

\subsection{Integration of Learned Components}

Machine learning techniques have enabled  leaps in  fundamental robotic tasks such as environment perception,  providing significant improvement   over traditional methods. There is also growing interest in applying learning techniques for planning, control, and general decision-making, for their ability to handle complex and dynamic scenarios. The latest emergence of large foundation models provides further opportunities for improving the understanding of complex multi-modal sensory data, enhancing decision-making and problem-solving capabilities, and facilitating human-robot interactions. 
However, the integration of machine learning  also brings significant, multi-faceted challenges to robotics, from the cost of collecting and labeling data, to explaining a component's decision. Among those challenges, providing assurances that the learned components will generalize and be robust to the intended physical environments is particularly crucial for safe and cost-effective robot deployments. Providing such assurances, however, is  extremely difficult in complex environments involving human and unstructured hazardous conditions. 
Assurance is further hampered by typically weak contracts between black-box learned components and other traditional components, and limited by the scaling of existing white-box analysis for learned models to overall integrated system states.

\subsection{Broken Scalability Tenets in   Robots' Software Pipelines}

The availability of high-quality, high-bandwidth, low-cost sensors like cameras and LiDARs has enabled robots to be more aware of and more responsive to their environments. 
The addition of such rich high-bandwidth sensors, however, is placing a significantly higher loads on robots' software data pipelines, from buses and memory to inter-process communication subsystems. These are eventually breaking the tenet of ``trivial communications delays'' between software components. 
This points to the need for a fundamental shift from current practices on how these systems are built, and more specifically to a requirement of new contracts and types of contracts between future components.
An orthogonal scalability tenet that is being challenged has to do with transferring the success from a single robot to multiple robots. Most successful  software pipelines have been designed for individual robots. 
However, successful deployment of an individual robot  does not easily scale to multiple robots, even if they are of the same kind. The deployment difference is not just quantitative, but rather qualitative, and includes issues ranging from higher-level planning, de-confliction, and avoidance, to communication, upgrades management, and monitoring and repair. Support to handle such dimensional scaling is lacking. 
 
\subsection{Diverse Siloed Development}
  
The diversity of development environments, processes and tools, software and hardware, assurance needs and regulations, and developers' backgrounds in robotics is overwhelming.  
Robotics development is performed in research, government, and commercial contexts. 
Many of these systems are safety-critical but to varying degrees, and they operate and are developed in disparate regulatory environments.
Commercial entities develop their own protocols, models, formats, and artifacts, often in closed-source contexts, to protect intellectual property.  
Researchers are more likely to share tools and artifacts, though these, too, are heterogeneous and often non-standardized. A lack of consensus on software engineering practices and tools in the robotics domain means that it can 
be difficult for others to adopt or extend those practices or resulting artifacts.
In contrast with more conventional software domains, roboticists also have particularly diverse backgrounds and education.
This is true of practicing engineers as well as students and researchers.  
Multidisciplinary engineers are not always sufficiently trained or experienced to make appropriate choices surrounding software development methodology.
This diversity challenges the unification of technology, abstractions, processes, and tools across the robotics ecosystem, and the aggregation of expertise. Robot systems knowledge and resources are thus distributed and siloed across diverse organizations with often incompatible incentives.

\section{Research Directions}

To address those challenges, the following  research directions are critical. 

\subsection{Modern Abstractions for Resources into Middlewares and Libraries}

\vspace{0.05in} \noindent \textit{Context.} 
Robotics software systems are typically structured as distributed systems, with components communicating via message passing or similar mechanisms, facilitated by middleware.  
Robotic middlewares and libraries provide specialized abstractions and off-the-shelf implementations of common functional blocks, and have become crucial in developing the software that drives robots.
Middlewares aim to simplify and standardize the integration of code written by different authors and organizations. They allow complex robot systems to be built as a collection of subsystems, often including both open-source and proprietary components.
In addition, middlewares accelerate scientific collaboration by simplifying the process of replacing individual subsystems.
As one example, a research group can iterate on the perception system of a mobile robot, while using pre-existing navigation and other subsystems, later distributing their work for research replication and comparative analysis.
Middlewares aim to reduce the friction of these technical integration challenges, promote code reuse and collaboration, and generally reduce unnecessary duplication of work.

Despite considerable progress, however, the design and implementation of a truly robust and flexible system for robotics messaging remains an open question. Robots continue to scale in terms of the number of components, layers, sensors, processors, and network types. 
This scaling challenges the assumptions underlying popular middlewares and libraries, such as the fact that communication is assumed to be ``fast enough to be reliable''.
Meanwhile, 
some modern sensors produce 100 times more data than a decade ago, wireless networks suffer from congestion and variability, and the significant performance penalties when naively traversing the many layers of caching in modern main and video memory systems.
These trends are limiting what can be achieved and slowing down development, making integration more difficult, and reuse more challenging.  Deploying software components to appropriate computing nodes is devolving into a manual, trial-and-error process. Above all, most roboticists do not wish to become middleware or deployment experts.
They simply want the underlying middleware plumbing to ``just work" so they can focus on robotics problems!

\vspace{0.05in} \noindent \textit{Vision.} 
We envision novel robotics middlewares and libraries that can handle the increased complexity of state-of-the-art robotics systems, with particular capability to reason about and manage resources.  These middlewares should support designing, configuring, and deploying integrated robotic systems involving complicated arrangements of sensors, actuators, computational resources (e.g., FPGAs, GPUs, CPUs), and networking resources (e.g., switches, routers, in-computer memory links), along a spectrum of different SWaP and deployment constraints (underwater, space, home, research, industry). Moreover, we envision capabilities for better understanding the computational, network, and hardware resources requirements necessary to run robotic software components, and their integration requirements.  

\vspace{0.05in} \noindent \textit{Research Agenda.} 
The research agenda must encompass the revision of fundamental abstractions with their corresponding explicit modeling APIs for accessing, understanding, configuring, and interfacing with system components and resources in the whole robotic stack.
Among those abstractions, communication is a key.
Explicitly modeling lossy links, QoS, latency, bandwidth, uncertainty, etc., as a part of the middleware to understand how the edges of the computational graph are grounded to the actual system is essential to supporting modern robotics use-cases such as mobile systems on wireless links, deployed underwater, in space, or even ad-hoc networks in a research lab.
This will also enable better diagnostics of underlying system components and support network links with non-perfect performance.
A second research direction is the development of new theories and techniques to automatically configure those abstractions to a particular context, similar to what is currently performed for sensor calibration.
A third research thread should aim to develop mechanisms to, given a system configuration and target scenario, characterize average and worst† case performance, and to provide recommendations to overcome limitations.

\vspace{0.05in} \noindent \textit{Specific Impact.} 
Any roboticist can describe difficulties they have had with the integration of a system, especially in handling the large volume of sensor data, responsiveness with actuation, and issues with bad network links.
First-class support of these concepts in the middleware will simplify the development process for researchers, industry integrators, and practitioners when it comes to developing new systems.
The ideal system will automatically and seamlessly handle a wide variety of network, computational, and traffic conditions for the vast majority of use cases, while still providing a clean and rational escalation path should human expertise and manual configuration be required.
The variety of robotics use-cases and environments makes the creation of such a general-purpose solution tremendously difficult but with a huge potential impact on virtually every real-world robotic system.
\subsection{Multi-View and Full Stack ADLs for Robotics}

\vspace{0.05in} \noindent \textit{Context.} 
The architectural complexity of software-intensive systems can be managed through different levels of abstractions and perspectives (known as an architectural \emph{view}). Those abstractions and views are facilitated by formal languages known as Architecture Description Languages (ADLs).
Intra-layer and single-view domain-specific ADLs have mostly sufficed for traditional software. 
Indeed, many elements of the robotics problem have corresponding modeling approaches and languages (e.g., for safety, architecture, kinematics, and more).  
However, the challenges of real-world robotics deployment and the leaky heterogeneous layers strain the capacity of existing modeling languages and approaches.  
There do not exist unifying approaches to compose these disparate models across their varying levels of abstraction, nor even agreement on what the appropriate abstraction levels might be. 
For the deployment of robots in the real world, it is necessary to enable the composition of robotics views across disciplines, domains, and system stack layers. This can be a unifying and enabling common ground across robotics development.

\vspace{0.05in} \noindent \textit{Vision.} 
We envision robotics-centric  
architecture description languages that span multiple layers and views, suitable for describing and analyzing the full stack of robotics software.  Such ADLs can overcome the gaps between areas of expertise and even disciplines, and provide automated support to reduce integration faults and development costs. The complexity of today's robotic systems has tipped the balance to make this investment worth it. 

\vspace{0.05in} \noindent \textit{Research Agenda.} 
Achieving that vision requires research into specification, architecture, design, synthesis, and analysis towards developing hierarchical architecture description languages (ADL) for robot system specification and design that accounts for the complementary and heterogeneous collection of views that manifest along the robotic layers. The hierarchy will have to be rich enough to represent a diverse set of robotic systems properties expressed through diverse views ranging, for example, from morphology, kinematics, and dynamics (e.g., URDF, kinematic trees, system of differential equations) to sensor properties and components behaviors  (e.g., non-linear probability distributions for accuracy and precision of sensors and actuators, confusion matrices for perception and action).
Such ADLs must also provide an ``inter-model API'' or set of connections, describing how views/models can be semantically related and translated in order to enable their broader consumption and composition. Understanding and refining the semantics of the views, how they translate to each other, what aspects are not translatable, and how to deal with those will constitute the main research challenge. A secondary research challenge is how to take advantage of those views to support development across the lifecycle. For example, such an ADL may enable automatic translation mechanisms from model to input for testing, or compilation technology that can automatically translate models into robotics code. 

\vspace{0.05in} \noindent \textit{Specific Impact.} 
The envisioned ADLs will increase developers' productivity by enabling, for example, faster design exploration (through the consistent combination, exploration,  and integrated understanding and composition of views), early implementation (automated synthesis of multiple partial views),  automated testing (by checking consistent behaviors across views), quicker debugging (through an integrating analysis of multiple views), and less expensive maintenance (through reusing models/views and better synchronization of changes).

\subsection{Processes, Models, Estimators, and Analyses for Humans and Robot Interactions}

\vspace{0.05in} \noindent \textit{Context.} 
Human-robot interaction (HRI) is poised to be a primary component of many near-future robotics systems.  Applications include service robotics for the home and business, co-robotics in industry and manufacturing, healthcare robotics,  autonomous vehicles for transportation, and human/robotic teams for security, defense, and disaster response. 
However, before robots are fully accepted in daily life tasks, they must act in a way that humans can trust, understand, predict, and accept.  To the extent that robots do not meet these expectations (or worse, are involved in accidents), humans will not accept their presence.  In close collaboration, meeting human expectations additionally requires robots that are able to model and predict human actions and reactions. We presently lack unified processes, models, estimators, and analyses to sustain those activities.  Worse, the human-in-the-loop is often left unconsidered into late in development. 

\vspace{0.05in} \noindent \textit{Vision.} 
We envision a robot development process that cost-effectively and continuously accounts for human motion, beliefs, desires, and intents, to produce systems that are trusted and accepted by humans. 
A subset of the systems produced by this process will themselves need to continuously account for human motion, beliefs, desires, and intents.

\vspace{0.05in} \noindent \textit{Research Agenda.} 
The research agenda focuses on the design and development of a unifying infrastructure for Human Robotic Interaction (HRI) that includes (1) models (particularly dynamic ones) that account for human motion, beliefs, desires, and intents, (2)  estimators of human state based on the models, (3) techniques to integrate and compose  extant diverse models and estimators with those generated by roboticists, and (4) data analysis mechanisms for multi-rate multi-modal distributed sensors, that are cognizant of noise, resource constraints, and privacy requirements.

There are many emerging independent and non-validated models of human behavior and interaction without clearly defined semantics for their understanding and composition.
The challenge lies in how to systematically define the full list of what must be modeled, how the current models address these factors,  and where the gaps are, and also in defining rich composing operators that can, in turn, render richer models for both developers and robotic systems. 
Software Engineering offers core competencies in tools and formalisms for modeling, estimating, analyzing, and testing complex sociotechnical systems, but lacks specialized tools for considering physical, embodied systems interacting with humans in the same space. 

These models can take many forms, tailored to specific applications. For instance, in addition to data or cognitive science-driven models, we foresee opportunities to develop robotics-specific user study tools to create human-in-the-loop models of human interaction. A technique common to HRI research is ``Wizard of Oz'' experiments, where users or bystanders interact with a robot that is controlled by a (possibly secret) external operator. A developer of a robotic system will need to test how a robot affects not just users, but other people existing in the same physical space. In software engineering, there are processes for identifying, tracking, and designing around impacted stakeholders for systems, including those that are not direct users. This agenda includes adapting these processes for robotics systems, allowing for testing for impact, even early on in development.

\vspace{0.05in} \noindent \textit{Specific Impact.} 
This research agenda promises to 
deliver new collaborative tools for collectively building and sharing HRI models to engender or improve robotic applications in mixed human-robot environments.  It promises to impact all aspects of software engineering of these systems, from specification to assurance and testing, rendering robotic systems that are more trusted and accepted by humans.

\subsection{Massive Simulation Eco-systems}

\vspace{0.05in} \noindent \textit{Context.} Simulators are crucial for robot system development. They are currently used, for example, for design exploration, prototyping, testing, benchmarking, and evaluating deployment readiness.  
That said, simulators are still limited along multiple directions. 
First, they are individually limited in their time, spatial, and dynamical scale, offering limited opportunities for more extensive deployments necessary to explore and validate sophisticated robot properties. 
This is especially problematic when simulating robots that operate in large, unstructured environments, such as agriculture, where it is difficult to model numerous plants and environmental features at sufficiently high fidelity. Such modeling challenges are compounded by environmental factors such as weather, blowing dust, and interactions with the robot system affecting its own environment. 
Second, rarer and novel sensing modalities, such as radar, thermal imagery, or neuromorphic sensors, lack high-quality open-source datasets to incorporate into simulation. The performance and behavior of these sensors is frequently highly coupled to a manufacturer's implementation or the complexities of the underlying environment (machine subsystems causing hot spots in thermal imagery, interior physical structures impact radar and sonar returns, neuromorphic sensors are difficult to realistically simulate in discrete event simulations, etc.), which also makes it difficult to share datasets and simulated sensor models between researchers. 
Due to these disconnects, success in simulation is only loosely connected to success in a real-world deployment, and many relevant physical interactions that could in principle be straightforwardly implemented remain impossible to simulate in practice.
All told, individual simulators are limited in their capacity to fully exercise any particular robotic deployment in simulation.  
And finally, each simulator operators as a standalone and closed component in the development process, and is not designed to work well with other simulators or tools. 
This separate usage not only increases development cost, but misses an opportunity to reduce the simulation reality gap collaboratively. They moreover do not share and do not provide incentives to share resources, nor collaboration mechanisms at scale with the community.

\vspace{0.05in} \noindent \textit{Vision.} We envision the capability to massively deploy robots across multiple \emph{collaborating} simulators that tackle different portions of the simulation-reality gap, including challenging simulated environments developed by multiple stakeholders, while sharing the simulated world with other robots to better explore and validate their capabilities before field deployment.
Excitingly, recent advances in technologies, including computer vision powered by modern data-driven machine learning, virtual reality, computer gaming, and physical simulation, are reaching a point of convergence; this vision is therefore timely. Furthermore, the success of such frameworks in other fields (i.e., Minecraft) provide guidelines on how to achieve similar massive deployments in this one. 

\vspace{0.05in} \noindent \textit{Research Agenda.}  The vision requires technical advances toward the development of collaborative, distributive, and modular simulation ecosystems that enable contributors to develop and integrate new environments, simulation capabilities, and robots (that may become part of the environment). 
Such ecosystems will have to support the integration of simulators specialized in different aspects of robotics (e.g., dynamics, fluids, sensed images) so they can build on each other, a marketplace for a rich diversity of environments (i.e., forest, roads, warehouse, kitchen, operating room), and the opportunity to deploy robots for long periods and in diverse spaces across those simulators and environments. 
Research will also be required to lower the adoption bar for ecosystems to accommodate the wide variety of protocols, models, formats, languages, simulation features, robots, sensing modalities, and environments used.  It will also be critical for the ecosystems to provide incentives, perhaps adopted from successful massive multiplayer games, for stakeholders to contribute to the collaborative effort. 

\vspace{0.05in} \noindent \textit{Specific Impact.}
The envisioned ecosystem will substantially reduce the cost of building more sophisticated simulators as their power can be composed. It will also improve the robustness of the simulated robotics systems as they will be tested in more comprehensive and diverse environments. Last, it will enable better benchmarking, as the shared worlds will provide comparative performance analysis, and potentially create competition and quantification for best performance.

\subsection{Multi-Dimensional Quality Assurance for Robots}

\vspace{0.05in} \noindent \textit{Context.}
Ensuring safety, security, and overall correctness for such robotic systems is essential for their adoption. 
However,  the field lacks specialized quality assurance techniques that account for the unique challenges of this domain. 
Specifying safety, security, and functional properties that can drive verification and fuzz testing, and defining mutation techniques and oracles to validate such properties are challenged by the  complexities of system and environments and their sources of uncertainty. 
Similarly, generating representative environments that can serve as testing inputs or verification environments is challenged by the sophistication required to mimic the real physical world. 
Additionally, while data-driven models (e.g., for perception, prediction, planning, and control) increasingly play a crucial role in many modern robots, quality assurance methods often rely on pre-defined rules and tests, and operate without considering adversarial assumptions. Thus, these methods are challenged by their ability to learn, adapt, and navigate in unpredictable and adversarial real-world scenarios.
Lastly, the status quo relies on the targeted verification and testing of critical components with simulations that approximate reality, and primarily on expensive field-testing that can only expose a limited set of behaviors.
Therefore, extant quality assurance techniques are ill-equipped to provide a holistic evaluation of a robotic system's capabilities in different dimensions.

\vspace{0.05in} \noindent \textit{Vision.} 
We envision a rich suite of quality assurance techniques catered to the needs of the robotics domain. These techniques would handle the heterogeneity of components, their interactions and unified behavior, and account for the complexity of the configuration spaces and adversarial threats, and the richness and uncertainties of the environments. They would also provide measures of the extent and success of the performed assurance activities.  
 
\vspace{0.05in} \noindent \textit{Research Agenda.} 
To accomplish this vision, the research agenda includes multiple complementary directions. First, we must develop new adequacy criteria that can determine the testing thoroughness, and identify testing weaknesses, as per the new dimensions introduced by robotics, from the real-world attributes to the components middleware interactions exercised. 
Second, powerful test generation approaches, such as fuzz and configuration testing, must be adapted to be driven by these new adequacy criteria (instead of the currently standard code coverage metrics). 
So, for example, fuzzing could be guided by the portions of the physical world covered; configuration testing could be cognizant of the hardware configurations utilized. 
Integrating those approaches with existing simulation frameworks is also important to enable their performance at scale in increasingly realistic environments.
Third, offline verification must be complemented by new run-time verification approaches that leverage existing middleware infrastructure and particularly check for emerging behaviors at the component level. 
Fourth, these run-time verification techniques must be augmented with compositional approaches that can integrate and build on the findings at the component level, even when the components are heterogeneous and sit across the system layers. 
Fifth, while data-driven models tend to perform well in specific situations or environments in which they were trained, real-world environments are dynamic and unpredictable. Therefore, there is a need for robot systems that can adapt to unforeseen situations, handle unexpected inputs, and maintain safe and reliable operations beyond their expected operating conditions. This can be achieved by learning robust models with diverse data collection and augmentation, and building methods that validate the output of models with additional contextual information.
Lastly, robotic QA requires the development of mixed-reality environments that can reduce simulation-reality gaps by providing mechanisms to integrate input from the real world and simulation.

\vspace{0.05in} \noindent \textit{Specific Impact.} Bringing quality assurance techniques to the robotics domain is central to the safe and secure real-world deployment of useful and complex robotics systems. Current approaches are not only limited but costly and time-consuming, so investigating techniques that can reduce that cost and resources are likely to accelerate system deployment.

\subsection{Evidence-based Processes and Assurances for Robotics Development}

\vspace{0.05in} \noindent \textit{Context.}  
As robots are rapidly becoming more prevalent across a wide range of applications, environments,  and development contexts.  
This evolution has resulted in highly specialized and often fragmented knowledge, experience, and expertise across the field of robotics when it comes to processes, assurance, hazards, and safety. Overall, there are limited evidence-driven processes,  and no agreed-upon frameworks that incorporate this knowledge and render it broadly accessible to the development and research communities.  Meanwhile,  successful practices and information go unshared. 
Among those processes and practices, assurance activities are critical: such activities must link evidence to claims of reliability, safety, and security through structured arguments that can be produced and reviewed algorithmically.
These activities range from  gathering and managing evidence that a robot meets its intended functions and does not present undesired behaviors, to building confidence about the readiness of a robot for an intended environment so that it can be certified for operation. The complexity of the physical environment, the heterogeneity of components and leaky layers, and the emerging behaviors make the alignment of evidence, behaviors, and environments particularly challenging in robotics. 
  
\vspace{0.05in} \noindent \textit{Vision.} 
We envision the establishment of a robotics-oriented development framework that includes process templates and a rich knowledge base that simultaneously addresses software-hardware-environment concerns across a  diversity of settings and provides a solid basis for assurance.
These templates should incorporate and share knowledge drawn from successful (and unsuccessful) previous projects and can be tailored to the specific needs of downstream organizations.   
The knowledge base will encompass assurance standards, knowledge books, repositories of hazards and relevant assurance cases, and mechanisms to query knowledge to meet a particular context and to compose pieces of diverse knowledge to assist in building assurance cases.

\vspace{0.05in} \noindent \textit{Research Agenda.} 
The research agenda to support our vision entails a wide-scale interdisciplinary collaboration with industry and research across heterogeneous domains. The goal is to collect and categorize necessary data about successful (and unsuccessful!) processes and assurance, and then analyze and synthesize process templates and knowledge from that data.  
Conducting these studies will require overcoming institutional inertia and commercial wariness to share `secret sauce'.   The research agenda should also incorporate mechanisms to query, analyze, adapt, compose, re-deploy, and refine the processes and grow the knowledge base.  
Last, the assurance research agenda in robotics should investigate how to better and continuously align and connect the complex system semantics, the engineering domain knowledge applied in design and development, and  the evidence collected in the form of testing results, formal verification proofs, analyses, and inspections.

\vspace{0.05in} \noindent \textit{Specific Impact.} Evidence-driven processes and first-class assurance activities will allow robotics development to scale with the complexity, safety, and variety of next-generation systems, and allow engineers of all stripes to contribute to robotics development while providing the required assurance guarantees. It will benefit industry, research labs, and government sponsors that have robotic development programs so they are more effective.

\subsection{Curricula to Prepare Future SE-Robotics Developers}

\vspace{0.05in} \noindent \textit{Context.} There is a disconnect between how we teach students robotics and how real robotics software development occurs in practice.  
Students pursuing a degree in robotics often come from a mechanical, electrical, or computer engineering background. While they understand the challenges of sensor noise and dynamics, they do not see themselves as professional software engineers, and often are not aware of the foundations of software engineering nor the best practices. When these students enter the industry, they lack many of the basic software engineering tools necessary for success. The nature of their industry role presents challenges to developing these skills, as software development may not their primary job, even though the software they write is critical to the correct operation of a robotic system.  On the other hand, students that are trained as software engineers often lack an understanding of the particularities of the robotic domain and their implications for software development. 
A compounding problem is that educational materials are often built, and rebuilt in a bespoke fashion on a per-institution basis. This is a short-minded choice that sacrifices learning federated,   open-source, rich development models that allow developers to share and reuse code, and more quickly and correctly build new robotics systems.   

\vspace{0.05in} \noindent \textit{Vision.} A future workforce that is ready to take on next-generation robot development challenges is predicated on college graduates with the foundations and skills at the intersection of robotics and software engineering to contribute to the whole robot software stack.  
This vision coincides with the up-swell of interest in robotic-related education, and the increasing demand for robotic system-related skills has sharpened the need for standardized and updated curricula. 

\vspace{0.05in} \noindent \textit{Research Agenda.} 
Achieving that vision will require a research agenda that establishes a baseline set of competencies and creates a common core set of resources for students from diverse technical educational backgrounds to become contributors to the software stack required by robotics. This will involve
 identifying  the core features of a Robotics Software Engineering (SE) curriculum including 
(1)  software engineering principles focused on unique robotics attributes (e.g., uncertainty management, changing environments, interactions with hardware, multiple views, large configuration spaces), 
(2) fundamental courses that crosscut the software stack and are specific to robotics like formal modeling and analysis, decision-making,  and classical robotics,  (3) cross-disciplinary courses that acclimate students to working with the breadth of engineering disciplines needed to develop robotic systems and develop specifications for the different components of a robotic system, and (4)  best practices for robotics software engineering in industry that must be acquired, from change management to build and deployment practices. 
In parallel, this effort must develop and package shared, reusable, expandable, materials and environments for students to create and run robot code, and identify an appropriate platform to deliver those.   Last,  mechanisms will have to be developed to maintain materials, in sync with evolving robotic code bases and tools.

\vspace{0.05in} \noindent \textit{Specific Impact.} 
Moving towards a shared, open-source model for education can help build a more consistent and shared understanding of the field, and lessen the load and maintenance barriers faced by educators. Using a federated model to build such content allows individual researchers to focus on their core competencies while simultaneously building a broader body of work that remains up to date. From a software engineering perspective, characterizing the unique aspects of the robotics domains will help students to specialize in their field of interest and produce educational materials that remain up to date with industry and the state of the art.
Meanwhile,  students seeking employment will benefit competitively from an education that is more attuned to the needs of the field and readier for cross-disciplinary collaborations. It also has the potential to accelerate the transition of research into the industry pipeline, as those students are better prepared to produce artifacts aligned with industry.

\newpage \section{Concluding Remarks}

This report summarizes the findings of the Workshop in Software Engineering for Robotics (2023). 
The report's main contributions are: (1) the identification of the unique and most pressing challenges associated with the software development of robotic systems, and (2) the first cohesive research agenda at the intersection of software engineering and robotics that must be undertaken to address those challenges. 

Beyond the findings, we would like to highlight the unifying consensus among the attendees at the workshop from 
 both the software engineering about the following:

\begin{itemize}

\item The workshop offered a tremendous synergistic opportunity across communities. There is no other venue where such interactions could have happened, and there is plenty of appetite from the attendees to make this a running series.  Thank you National Science Foundation for enabling it!

\item    If successful, the proposed agenda has the potential to dramatically improve
how the software stack of robot systems is developed, enhancing overall performance, safely enabling new capabilities, and lowering cost, while 
impacting the robotic industry, the research labs, the workforce produced by universities, and the end users consuming robots and robot services. 

\item We are at a significant juncture, with urgent needs and significant opportunities, for domain-specific software development techniques, processes, and tools to not just maintain but also to accelerate and further enable the robotic revolution.  Funding for these opportunities is insufficient and scattered,  leading much of the required science and engineering to fall through the cracks of existing programs. 

\end{itemize}

\newpage
\section*{Workshop Agenda}
\label{sec:agenda}

This was a working workshop that tapped into the diverse experiences and expertise of the participants. We scheduled four framing talks to help highlight the different dimensions of the problem and the solution space. 
However, most of the schedule was allocated to hands-on activities designed by the organizers (inspired in part by the activities used by groups like Knowinnovation).

\begin{table}[h]
{\small
\begin{tabular}{lm{32em}}
\toprule \\
Day 1 - October 5th \\\midrule
8:45 - 9:00 &  Welcome \\
9:00 - 10:30 & Ideation I   \\
10:30 - 11:00 & Break and networking\\
11:00 - 12:00 & Framing talks I:\\
& - On writing robot software with ROS 1 and ROS 2: Some Lessons Learned, Morgan Quigley, Chief Architect at Open Robotics\\
& - Languages and Architectures for Guiding Long-lived Systems, Brian Williams, Professor at MIT\\
12:00 - 12:30 & Prioritizing best ideas\\
12:30 - 13:30 & Lunch and networking\\
13:30 - 14:30 & Ideation II\\
14:30 - 15:00 & Prioritizing best ideas\\
15:00 - 15:30 & Break and networking\\
15:30 - 16:45 & Framing talks II: \\
& - Towards a Design Flow for Verified AI-Based Autonomy, Sanjit Seshia, Professor at UC Berkeley \\
& - Robot Disintegration: How Common Robot Architectures Work     Against Successful Robot Products,  Todd Pack, Principal Robotics Engineer at Neo Cybernetica \\
\toprule
Day 2 - October 6th \\
\midrule
8:45 - 9:00 & Recap of Day 1 \\
9:00 - 10:30 & Theme synthesis and report I\\
10:30 - 11:00 & Break and networking\\
11:00 - 12:30 & Theme synthesis and report II\\
12:30 - 13:30 & Lunch and networking\\
13:30 - 15:00 & Theme refinement and report\\
15:00 - 15:30 & Closing\\
\bottomrule
\end{tabular}}
\end{table}

\subsection{Day 1: Ideation and Prioritization}

The first day, the hands-on activities alternated between Ideation with  Prioritization sessions.  

\vspace{0.05in} \noindent \textit{Ideation.} 
The organizers assigned participants to tables, with a goal of distributing the software engineering experts and robotics experts roughly evenly.  For the second session, to encourage mixing while maintaining some cohesion, the software engineering researchers moved as a group to new (designated) tables. 
Each table was provided with a large quantities of sticky notes.  For 10 minutes,  participants were asked to write as many ideas as they could generate, following the prompt ``Would be great to develop X to enable Y.''
Each table then spent 30 minutes discussing all ideas, giving feedback, and refining.  
At the end of the ideation sessions, participants  placed the stickies on a large sheet of paper taped to the wall of the room.  Because of the volume of ideas generated during the first session, for the second session, each table was asked to pick their 3--5 ``best'' ideas for contribution to the wall.  

Before the workshop, the organizers had sent a brief survey to the participants about their interests.  The most popular topics were projected at the front of the room to inspire the session: 
(1) Certification and assurance robot hw/sw 
(2) PL and abstractions for complex state, uncertainty, variability
(3) Integration/evolution of physical/cyber components
(4) AI in robotics, composition with learned components 
(5) HRI and impact on design, assurance, analysis
(6) Deployments and long term maintenance
(7) Education
(8) Other.

\vspace{0.05in} \noindent \textit{Prioritization.} 
For the first prioritization session, participants were encouraged to mingle and inspect the ideas on the wall.  The goal of the first session was to prioritize or identify the best ideas; signs were affixed to indicate ``hot'' ideas were to move right; ``cold'' ideas, left.  The participants were to discuss the ideas and, if two people agreed, could move an idea closer to ``hot''.  Only one person (without discussion) was required to move an idea closer to ``cold'' While this activity did encourage discussion, participants and organizers observed that it did not always lead to the desired prioritization.  Participants did not often move ideas ``down''.  Additionally, participants expressed a desire to cluster the ideas, as there was clear repetition and commonalities between them.
Therefore, for the second session, participants were additionally encouraged to work together to cluster the ideas by theme or topic. Each participant was also encouraged to mark up to three sticky notes with a dot to indicate a ``vote'' for that idea.  
\footnote{For future workshops, the activity might benefit from (1) encouraging tables to produce a smaller set of 5-7 (or 3-5) ideas for both sessions, and (2) making the clustering approach a primary objective of prioritization. A key goal of prioritization is to encourage discussion; a similar approach for clustering could allow two ideas to be ``clustered'' (or rewritten) if two participants agreed, but ``unclustered'' by only one.}

\subsection{Day 2: Idea Synthesis and Reporting}

Before the second day, the organizers collected and documented the clustered ideas by theme: (1) Assurance, (2) education, (3) HRI, (4) middleware, (5) modeling, (6) process, (7) simulation, (8) testing, and (9) other.  The participants then selected their own table by topic, and were encouraged to change tables/topics between sessions as desired. 

\vspace{0.05in} \noindent \textit{Theme synthesis and feedback.} 
Tables were asked to appoint a scribe, and then review the clustered sticky notes associated with their table's topic theme, discuss, and prepare a single slide (collaborating on a shared Google Slides presentation) synthesizing one or more of the best ideas from the stickies.  
Each slide included: (1) Working title + participants, (2) What is the potential impact towards robotics? (3) What is the potential impact towards SE? (4) What is novel? and (5) What’s been stopping us from doing this up until now?

Each group then presented their favorite idea to the whole workshop. The audience prepared feedback, again using sticky notes.  Each sticky note was encouraged to include (1) Pluses (+): What’s good about the idea the way you heard it?  (2) Potentials (P): What ideas came to you while listening ? and (3) Concerns (C): phrased as challenges. How to …? What might be all the…? In what ways might …?  The collected feedback was added to the slides by each group's scribe.
These sessions worked well; however, a future event should allocate more than 30 minutes to this activity.

\vspace{0.05in} \noindent \textit{Theme refinement.} 
For the last session, each table reviewed slides related to their themes and worked together to refine one or more ideas into a short written report, in a Google Document, answering the questions
(1) What are you trying to do?
(2) Why now?
(3) Why it's important?
(4) What are the barriers and challenges to overcome?
(5) How might this turbo-charge robotics ? SE?
and (6) If this is successful who will care?  These reports were collected in a shared Drive, and, along with the slides, formed the basis of the initial drafts of this overall report.

\end{document}